\documentclass[10pt,twocolumn,letterpaper]{article}

\usepackage[dvipsnames,table]{xcolor} 

\usepackage[pagenumbers]{iccv}

\usepackage{graphicx}
\usepackage{amsmath}
\usepackage{amssymb}
\usepackage{makecell}
\usepackage{multirow}
\usepackage{mathtools}
\usepackage{stmaryrd}
\usepackage{hhline}
\usepackage{tabularray}
\usepackage{booktabs}
\usepackage{placeins}
\usepackage{subfiles}
\usepackage{stfloats}
\usepackage{caption}
\usepackage{cuted}

\usepackage{tikz}
\usepackage{forest}
\useforestlibrary{edges}
\usetikzlibrary{calc, shadows, trees}
\usepackage{mciteplus}
\newsavebox{\myTaxonomyBox}
\usepackage[sectionbib]{chapterbib}

\usepackage[most]{tcolorbox}
\usepackage[percent]{overpic}
\usepackage{csvsimple}
\usepackage{adjustbox}
\usepackage[ddmmyyyy]{datetime}
\usepackage{pdflscape}
\usepackage{rotating} 
\usepackage{standalone}
\usepackage{afterpage}

\usepackage{ltablex}   
\usepackage{supertabular}
\keepXColumns
\usepackage{ragged2e}

\newcolumntype{C}[1]{>{\centering\arraybackslash}m{#1}}

\definecolor{grapefruit}{HTML}{DA4453}
\definecolor{bittersweet}{HTML}{E9573F}
\definecolor{sunflower}{HTML}{F6BB42}
\definecolor{grass}{HTML}{8CC152}
\definecolor{mint}{HTML}{37BC9B}
\definecolor{aqua}{HTML}{3BAFDA}
\definecolor{bluejeans}{HTML}{4A89DC}
\definecolor{lavender}{HTML}{967ADC}
\definecolor{pinkrose}{HTML}{D770AD}
\definecolor{grey}{HTML}{ECEEDA}
\definecolor{magenta}{HTML}{E377C2}
\definecolor{orange}{HTML}{FF7F0E}
\definecolor{edgesgrey}{rgb}{0.7,0.7,0.7} 
\definecolor{hidden-draw}{rgb}{0.7,0.7,0.7} 
\definecolor{iccvblue}{rgb}{0.21,0.49,0.74}

\definecolor{preproduction}{HTML}{34A853} 
\definecolor{production}{HTML}{FBBC04}   
\definecolor{postproduction}{HTML}{4285F4} 
\definecolor{redcolor}{HTML}{DA4453}      
\definecolor{bluecolor}{HTML}{4A89DC}     

\newtcbox{\xmybox}[1][red]{leftright skip=0.05cm, tcbox width=forced center, width=0.6cm, on line,
arc=7pt,colback=#1!10!white,colframe=#1!50!black,
before upper={\rule[-3pt]{0pt}{10pt}},boxrule=1pt,
boxsep=0pt,left=6pt,right=6pt,top=2pt,bottom=2pt}

\newtcbox{\xmyboxsquare}[1][red]{leftright skip=0.05cm, tcbox width=forced center, width=0.6cm, on line,
arc=0pt,colback=#1!10!white,colframe=#1!50!black,
before upper={\rule[-3pt]{0pt}{10pt}},boxrule=1pt,
boxsep=0pt,left=6pt,right=6pt,top=2pt,bottom=2pt}

\tikzset{
  my-box/.style={
    rectangle,
    rounded corners=3pt,
    minimum height=2em,
    minimum width=5em,
    inner sep=4pt,
    align=center,
    draw,
    line width=0.8pt,
    drop shadow={opacity=.5, shadow xshift=0.5mm, shadow yshift=-0.5mm}
  },
}

\usepackage[pagebackref,breaklinks,colorlinks,allcolors=iccvblue]{hyperref}

\usepackage{lipsum}

\title{A Survey on Long-Video Storytelling Generation: \\Architectures, Consistency, and Cinematic Quality}

\author{%
  \textbf{Mohamed Elmoghany}$^{1}$,
  \textbf{Ryan Rossi}$^{1}$,
  \textbf{Seunghyun Yoon}$^{1}$,
  \textbf{Subhojyoti Mukherjee}$^{1}$,\\
  \textbf{Eslam Bakr}$^{2}$,
  \textbf{Puneet Mathur}$^{1}$,
  \textbf{Gang Wu}$^{1}$,
  \textbf{Viet Dac Lai}$^{1}$,
  \textbf{Nedim Lipka}$^{1}$,\\
  \textbf{Ruiyi Zhang}$^{1}$,
  \textbf{Varun Manjunatha}$^{1}$,
  \textbf{Chien Nguyen}$^{1,3}$,
  \textbf{Daksh Dangi}$^{4}$,\\
  \textbf{Abel Salinas}$^{5}$,
  \textbf{Mohammad Taesiri}$^{4}$,
  \textbf{Hongjie Chen}$^{6}$,
  \textbf{Xiaolei Huang}$^{7}$,\\
  \textbf{Joe Barrow}$^{11}$,
  \textbf{Nesreen Ahmed}$^{8}$,
  \textbf{Hoda Eldardiry}$^{9}$,
  \textbf{Namyong Park}$^{12}$,
  \textbf{Yu Wang}$^{3}$,\\
  \textbf{Jaemin Cho}$^{15}$,
  \textbf{Anh Totti Nguyen}$^{14}$,
  \textbf{Zhengzhong Tu}$^{10}$,
  \textbf{Thien Nguyen}$^{1}$,
  \textbf{Dinesh Manocha}$^{13}$,\\
  \textbf{Mohamed Elhoseiny}$^{2}$,
  \textbf{Franck Dernoncourt}$^{1}$\\[1ex]
  $^{1}$Adobe Research \quad
  $^{2}$KAUST \quad
  $^{3}$University of Oregon \quad
  $^{4}$Independent Researcher \quad\\
  $^{5}$University of Southern California \quad
  $^{6}$Dolby Labs \quad
  $^{7}$University of Memphis \quad
  $^{8}$Cisco \quad\\
  $^{9}$Virginia Tech \quad
  $^{10}$Texas A\&M University \quad
  $^{11}$Pattern Data \quad
  $^{12}$Meta AI \quad\\
  $^{13}$University of Maryland, College Park \quad
  $^{14}$Auburn University \quad
  $^{15}$UNC Chapel Hill
}

\begin{document}
\maketitle

\begin{abstract}
 Despite the significant progress that has been made in video generative models, existing state-of-the-art methods can only produce videos lasting 5-16 seconds, often labeled “long-form videos”. Furthermore, videos exceeding 16 seconds struggle to maintain consistent character appearances and scene layouts throughout the narrative. In particular, multi-subject long videos still fail to preserve character consistency and motion coherence. While some methods can generate videos up to 150 seconds long, they often suffer from frame redundancy and low temporal diversity. Recent work has attempted to produce long-form videos featuring multiple characters, narrative coherence, and high-fidelity detail. We comprehensively studied 32 papers on video generation to identify key architectural components and training strategies that consistently yield these qualities. We also construct a comprehensive novel taxonomy of existing methods and present comparative tables that categorize papers by their architectural designs and performance characteristics.
\end{abstract}
    
\tikzstyle{my-box}=[
    rectangle,
    draw=hidden-draw,
    rounded corners,
    text opacity=1,
    minimum height=2em,
    minimum width=5em,
    inner sep=2pt,
    align=center,
    fill opacity=.5,
    line width=0.8pt,
]
\tikzstyle{leaf}=[my-box, minimum height=2em,
    fill=white!80, text=black, align=left,font=\normalsize,
    inner xsep=2pt,
    inner ysep=4pt,
    line width=0.8pt,
]
\begin{figure*}[!t]
    \centering
    \resizebox{0.85\textwidth}{!}{
        \begin{forest}
            forked edges,
            for tree={
                grow=east,
                reversed=true,
                anchor=base west,
                parent anchor=east,
                child anchor=west,
                base=left,
                font=\large,
                rectangle,
                draw=hidden-draw,
                rounded corners,
                align=left,
                minimum width=6em,
                edge+={edgesgrey, line width=1pt},
                s sep=20pt,
                inner xsep=5pt,
                inner ysep=5pt,
                line width=0.8pt,
                ver/.style={rotate=90, child anchor=north, parent anchor=south, anchor=center},
            },
            where level=1{text width=13em,font=\normalsize,}{},
            where level=2{text width=10em,font=\normalsize,}{},
            where level=3{text width=8.2em,font=\normalsize,}{},
            where level=4{text width=8.2em,font=\normalsize,}{},
            [
                \textbf{Long-Video}\\\textbf{Generation}\\~(\S \ref{sec:taxonomy})
                        , my-box, text width=11em
                    [
                    \textcolor{grapefruit}{\textbf{Keyframes-to-Video}}\\~(\S \ref{sec:arch-keyframe})
                    [
                        \textcolor{bluecolor}{\quad MovieAgent}~\cite{movieagent2025}{, }
                        \textcolor{bluecolor}{\quad StoryDiffusion}~\cite{zhou2024storydiffusion}{, }\\
                        \textcolor{bluecolor}{\quad KeyVID}~\cite{wang2025keyvid}{, }
                        \textcolor{bluecolor}{\quad I2VGen-XL}~\cite{zhang2023i2vgenxl}{, }\\
                        \textcolor{bluecolor}{\quad AutoStory}~\cite{liu2023autostory}{, }
                        \textcolor{bluecolor}{\quad StableVideoDiffusion}~\cite{ho2023svd}{ }
                        , leaf, text width=22em
                    ]
                ]
                [
                    \textcolor{orange}{\textbf{Discrete Temporal Chunks}}\\~(\S \ref{sec:arch-chunks})
                    [
                        \textcolor{bluecolor}{\quad MAGI-1}~\cite{teng2025magi1}{, }
                        \textcolor{bluecolor}{\quad Ca2-VDM}~\cite{gao2024ca2vdm}{, }
                        \textcolor{bluecolor}{\quad SEINE}~\cite{chen2023seine}{ }
                        , leaf, text width=22em
                    ]
                ]
                [
                    \textcolor{magenta}{\textbf{High Compression}}\\~(\S \ref{sec:arch-compression})
                    [
                        \textcolor{bluecolor}{\quad FramePack}~\cite{zhang2025framepack}{, }
                        \textcolor{bluecolor}{\quad LTX-Video}~\cite{hacohen2025ltxvideo}{, }\\
                        \textcolor{bluecolor}{\quad Open-Sora Plan}~\cite{lin2024opensoraplan}{, }
                        \textcolor{bluecolor}{\quad PyramidFlow}~\cite{jin2024pyramidflow}{ }
                        , leaf, text width=22em
                    ]
                ]
                [ 
                    \textcolor{grass}{\textbf{Flattened 3D Space-Time}}\\\textcolor{grass}{\textbf{(One-Shot)}}\\~(\S \ref{sec:arch-one-shot})
                    [
                        \textcolor{grass}{\textbf{Foundational}}\\\textcolor{grass}{~(\S \ref{sec:arch-found})}
                        [
                            \textcolor{bluecolor}{\quad HunyuanVideo}~\cite{kong2024hunyuanvideo}{, }
                            \textcolor{bluecolor}{\quad WAN2.1}~\cite{wan2025}{, }\\
                            \textcolor{bluecolor}{\quad Open-Sora2.0}~\cite{peng2025opensora2}{, }
                            \textcolor{bluecolor}{\quad VACE}~\cite{vace2025}{, }\\
                            \textcolor{bluecolor}{\quad MAGVIT-v2}~\cite{yu2023magvitv2}{, }
                            \textcolor{bluecolor}{\quad Lumiere}~\cite{bartal2024lumiere}{, }\\
                            \textcolor{bluecolor}{\quad VideoDiffusionModels}~\cite{ho2022vdm}{ }
                            , leaf, text width=20em
                        ]
                    ]
                    [
                        \textcolor{grass}{\textbf{Single-Subject}}\\\textcolor{grass}{\textbf{Personalization}}\\~(\S \ref{sec:arch-single})
                        [
                            \textcolor{bluecolor}{\quad MovieGen}~\cite{polyak2024moviegen}{, }
                            \textcolor{bluecolor}{\quad Phantom}~\cite{liu2025phantom}{, }\\
                            \textcolor{bluecolor}{\quad CogVideoX}~\cite{yang2024cogvideox}{, }
                            \textcolor{bluecolor}{\quad DreamVideo}~\cite{gao2023dreamvideo}{, }\\
                            \textcolor{bluecolor}{\quad AnimateDiff}~\cite{gu2023animatediff}{ }
                                , leaf, text width=20em
                        ]
                    ]
                    [
                        \textcolor{grass}{\textbf{Multi-Subject}}\\\textcolor{grass}{\textbf{Personalization}}\\~(\S \ref{sec:arch-one-multi})
                        [
                            \textcolor{bluecolor}{\quad VideoAlchemist}~\cite{chen2025videoalchemist}{, }
                            \textcolor{bluecolor}{\quad SkyReels-v2}~\cite{chen2025skyreelsv2}{, }\\
                            \textcolor{bluecolor}{\quad ConceptMaster}~\cite{huang2025conceptmaster}{, }
                            \textcolor{bluecolor}{\quad HunyuanCustom}~\cite{hu2025hunyuancustom}{ }
                                , leaf, text width=20em
                        ]
                    ]
                    [
                        \textcolor{grass}{\textbf{Multi-Shot Narrative}}\\\textcolor{grass}{\textbf{Planning}}\\~(\S \ref{sec:arch-narrative})
                        [
                            \textcolor{bluecolor}{\quad Seedance}~\cite{gao2025seedance}{, }
                            \textcolor{bluecolor}{\quad StepVideo}~\cite{huang2025stepvideo}{, }\\
                            \textcolor{bluecolor}{\quad SkyReels-v2}~\cite{chen2025skyreelsv2}{ }
                                , leaf, text width=20em
                        ]
                    ]
                ]
                [
                    \textcolor{mint}{\textbf{Token-Stream}}\\\textcolor{mint}{\textbf{Autoregressive-Token}}\\~(\S \ref{sec:arch-token})
                    [
                        \textcolor{bluecolor}{\quad Loong}~\cite{wang2024loong}{, }
                        \textcolor{bluecolor}{\quad VideoPoet}~\cite{singer2023videopoet}{ }
                        , leaf, text width=22em
                    ]
                ]
                [
                    \textcolor{lavender}{\textbf{Closed Source}}\\~(\S \ref{sec:arch-closed})
                    [ 
                        \textcolor{bluecolor}{\quad Veo3}~\cite{deepmind2025veo3}{, }
                        \textcolor{bluecolor}{\quad Pika 2.2}~\cite{pika1p5}{, }
                        \textcolor{bluecolor}{\quad Kling2.1}~\cite{kling2p1}{, }\\
                        \textcolor{bluecolor}{\quad Seedance}~\cite{gao2025seedance}{, }
                        \textcolor{bluecolor}{\quad MovieGen}~\cite{polyak2024moviegen}{, }
                        \textcolor{bluecolor}{\quad Sora}~\cite{openai2024sora}{, }\\
                        \textcolor{bluecolor}{\quad Minimax Hailuo-02}~\cite{hailuo2025}{, }
                        \textcolor{bluecolor}{\quad RunWay Gen3}~\cite{runwaygen3}{ }
                        , leaf, text width=22em
                    ]
                ]
            ]
        \end{forest}
    }
    \caption{Architectural taxonomy of long-video generation methods. These trees were selected because together they span the key axes of temporal decomposition, compression, personalization scope, narrative structure, and openness that govern modern long-video synthesis.  \textcolor{grapefruit}{\textbf{(a) Keyframes-to-Video:}} two-stage generating images as keyframes followed by motion pipelines that scale to minute-long clips \textcolor{orange}{\textbf{(b) Discrete Temporal Chunks:}} constant-memory, parallel decoding of N-frame blocks \textcolor{magenta}{\textbf{(c) High Compression:}} heavy latent down-sampling for real-time, low-power inference \textcolor{grass}{\textbf{(d) Flattened 3D One-Shot:}} end-to-end full-tensor synthesis: 1. Foundational: joint spatiotemporal prior for fixed-length clips; 2. Single-Subject: adapters for identity consistency and faithful individual likeness; 3. Multi-Subject: dedicated per-entity fusion modules for coherent multi-character scenes; 4. Multi-Shot: shot-level segmentation for structured scene planning \textcolor{mint}{\textbf{(e) Token-Stream Autoregressive:}} unified text-video token decoding with maximal modality flexibility \textcolor{lavender}{\textbf{(f) Closed-Source:}} proprietary systems that set the current quality ceiling.
    }
    \label{fig:main-taxonomy}
\end{figure*}
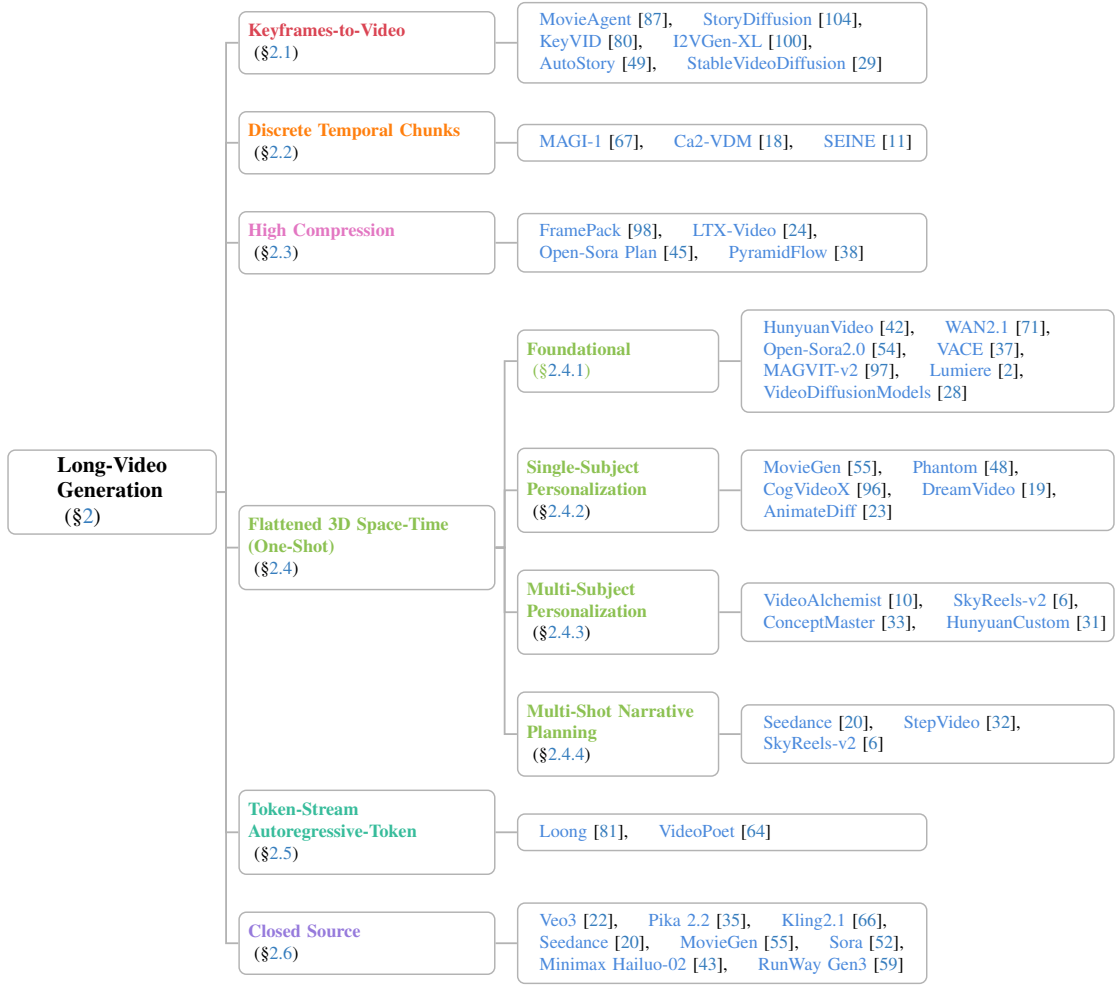

\section{Introduction}
\label{sec:intro}

The advent of diffusion-based models DDPM \cite{ddpm2020} and DDIM \cite{ddim2020}, together with recent advances in large language models \cite{brown2020language}, has laid the foundation for cinematic long-form video generation and AI-driven content creation. High-quality, realistic video generation now underpins applications in education \cite{wang2023text2video_education}, marketing \cite{zhang2023diffusion_marketing}, autonomous driving \cite{wen2024panacea_autonomous_driving}, gaming \cite{muller2020gamegan}, entertainment \cite{kim2023cinediff}, robotic learning \cite{james2022sim2real}, medicine \cite{xu2020surgical_video_synthesis}, and virtual reality \cite{yang2021vr360}. These modeling developments produce large volumes of synthetic video data that benefit all of the aforementioned domains. Such data can support educational content, drive task-specific applications, or facilitate the training of next-generation machine learning models.

Video generation poses greater challenges than text or image generation. It incorporates temporal complexity alongside spatial complexity, ensuring frames' consistency, which is a key distinction from image generation. Consequently, beyond 16 seconds, autoregressive methods accumulate quality degradation, leading to inconsistencies, visual artifacts, or other perceptual errors \cite{xing2023survey}. Another limitation is the high memory requirements and computational resources required for video generation, as in the spatiotemporal attention \cite{wang2018nonlocal,duke2021sstvos,bertasius2021timesformer}.

Video datasets that can be used commercially are very limited, which further hampers progress. Most public datasets require commercial licenses i.e. MovieBench \cite{moviebench2025}, Koala-36M \cite{wang2024koala36m}, CelebV-HQ \cite{zhu2022celebvhq}, Panda-70M \cite{chen2024panda70m}, HD-VG-130M \cite{videofactory} and MiraData \cite{ju2024miradata}, impeding industry-driven innovation. Long-form video generation demands realistic clips with detailed annotations to capture full narrative scenes. However, existing open-source datasets typically contain only a few seconds of footage. Moreover, crucial metadata such as shot type, camera motion, character emotions, background context, and action labels are rarely provided, necessitating custom dataset creation and curation. Richly annotated datasets, as in MovieBench \cite{moviebench2025}, include details about the background, characters, camera shot type, and camera motion. All of these annotation details of the dataset would improve text-to-video alignment and prompt adherence as per their benchmark results. In addition, the same dataset includes high-quality long videos, leading to the capability of generating longer videos.

Maintaining character appearance consistency over time poses a challenge \cite{tulyakov2018mocogan}. Camera-relative character motion alters scale and density, potentially disrupting visual continuity \cite{brooks2022generating}. Ensuring temporal scene consistency during character movement is also critical \cite{yan2022temporally}. Supporting multiple characters while preserving consistency is even more demanding \cite{lin2023videodirectorgpt}. Abrupt scene transitions hinder coherent narrative generation \cite{wu2025mind}. Enforcing physical plausibility and accurate interactions further complicates video synthesis \cite{yang2025towards}.

In this survey, we highlight the core architectural elements and training methodologies that reliably address the primary obstacles in video generation. We introduce a foundational framework that identifies and recommends a set of novel, high-potential components whose integration can overcome these challenges and enable the synthesis of longer, coherent, and visually compelling videos. 


Our contributions in this paper are summarized as:
\begin{enumerate}
  \item \textbf{Comprehensive Taxonomy and Architectural Backbone:} We introduce a taxonomy tree in Section~\ref{sec:taxonomy} that organizes video‐generation methods by their primary architectural focus, and Table~\ref{tab:backbone1} presents a detailed comparison of the core components used in state‐of‐the‐art models.

  \item \textbf{Architectural Advances:} We survey emerging design patterns in video generation—covering training objectives, backbone networks, text encoders, VAE variants, positional embeddings, and other core modules. Section~\ref{sec:architecture} distills these components to guide the next generation of foundational video‐diffusion models.

  \item \textbf{Datasets and Evaluation Metrics:} We identify underexplored cinematic and long‐form video datasets with high potential, and catalog the evaluation metrics recently adopted by the community (see Appendix).
\end{enumerate}
\section{Long Video Generation Architectural Styles}
\label{sec:taxonomy}
We organize video‐generation research into 6 distinct taxonomies (Fig.~\ref{fig:main-taxonomy}), and for each taxonomy tree we provide a detailed analysis along with principled guidelines for selecting the most suitable architectural paradigm. For example, a category like Keyframes-to-Video focuses on generating the video through keyframes images. While a category like discrete temporal chunks, focuses more on dividing the generation into multiple smaller generations stitched together.

\subsection{Keyframes-to-Video}
\label{sec:arch-keyframe}
Unlike one‐shot text‐to‐video methods, several recent works~\cite{chen2025hunyuanvideoavatar,movieagent2025,zhou2024storydiffusion,zhang2023i2vgenxl,liu2023autostory,ho2022vdm} employ a two‐stage generation pipeline. For example, KeyVID~\cite{wang2025keyvid}, unlike others, uses audio-to-video modality. It first partitions the audio stream into audio latent keyframe segments and then synthesizes video for each segment. Likewise, StoryDiffusion~\cite{zhou2024storydiffusion} decomposes the narrative text into a series of sub‐prompts, generates a keyframe image per sub‐prompt, and uses a motion‐prediction module to interpolate these keyframes into a continuous video sequence. This two‐stage paradigm enables scalable long‐duration video synthesis by producing keyframes, followed by motion interpolation or generation, and stitching the resulting segments into arbitrarily long videos. This methodology ensures the consistency of the whole video semantically. However, the sequential nature of keyframe and video synthesis incurs extra latency compared to end‐to‐end approaches, and requires distinct text‐to‐image and text‐to‐video models when a single model does not natively support both modalities.


\subsection{Discrete Temporal Chunks}
\label{sec:arch-chunks}
In this approach, a video is partitioned into disjoint temporal chunks of N frames (e.g., 8, 16, or 25). Each chunk is generated independently and then concatenated to reconstruct the full sequence. By capping memory usage at the chunk level, this scheme significantly reduces peak GPU requirements and naturally supports chunks parallel processing. The primary drawback is the potential for artifacts at chunk boundaries. Several recent studies employ this chunk-based paradigm \cite{teng2025magi1, gao2024ca2vdm, chen2023seine}. For example, MAGI-1 \cite{teng2025magi1} partitions videos into 24-frame segments and holistically denoises each one. The denoised output then conditions the subsequent segment, with four segments processed concurrently. CA2-VDM \cite{gao2024ca2vdm} further demonstrates that such chunked training schedules often require additional epochs to learn the diverse chunked boundaries across segments.


\subsection{High Compression}
\label{sec:arch-compression}
Most existing video-generation models demand high-end GPUs at inference. To address this, several recent works~\cite{zhang2025framepack, hacohen2025ltxvideo, lin2024opensoraplan, jin2024pyramidflow} have aggressively reduced model size and parameter count. LTXVideo~\cite{hacohen2025ltxvideo} introduces Video-VAE, a variational autoencoder that compresses spatiotemporal dimensions by 192× into a 128-channel latent tensor without patchification, drastically reducing token count and enabling low-latency inference; however, this level of compression sacrifices fine-grained texture details and subtle motions, and can introduce artifacts in regions of rapid movement. FramePack~\cite{zhang2025framepack} enforces a fixed context length via relative temporal weighting, assigning 50\% to the most recent frame, 25\% to the prior, 12.5\% to the next, etc. This supports efficient extension across arbitrary durations but often yields outputs with limited diversity in background and motion.

\subsection{Flattened 3D Space-Time (One-Shot)}
\label{sec:arch-one-shot}
The flattened 3D space–time one-shot approach regards the entire video as a single spatiotemporal tensor and synthesizes it in a single forward pass. By folding the temporal axis into a unified spatial latent and applying 3D convolutional diffusion-transformer blocks, these models jointly capture cross-frame dependencies to produce high-fidelity clips of fixed duration as in ~\cite{ho2022vdm, yu2023magvitv2}. However, this end-to-end formulation imposes heavy GPU requirements, which in turn limit achievable clip length and spatial resolution~\cite{wan2025}. Crucially, it ensures semantic coherence across frames while effectively modeling long-range motion correlations. Most of the video generation models are using one-shot techniques. However, the one-shot models differ in their focus as in WAN2.1 \cite{wan2025} aiming to build a foundational model. On the other side, phantom \cite{liu2025phantom} focuses more on subject-personalization. 
\subsubsection{Foundational One-Shot}
\label{sec:arch-found}
Foundational one-shot models formalize the essential blueprint for treating the entire video as a unified spatiotemporal tensor and generating it in a single pass. They learn a joint prior over the full video tensor via 3D UNet Diffusion, DiT, or MM-DiT backbones ~\cite{unet2015, dit2023, flowmatching2023, flowmatching2024}, enabling one-shot synthesis of videos. That is beside using a 3D VAE that compresses each clip and converts it into a dense latent grid. Papers like ~\cite{kong2024hunyuanvideo, wan2025, peng2025opensora2, vace2025, yu2023magvitv2, bartal2024lumiere, ho2022vdm} focus on creating Foundational models. For instance, WAN2.1 ~\cite{wan2025} uses a single DiT instead of using MM-DiT; however it complements by using a cross-attention module. While HunyuanVideo ~\cite{kong2024hunyuanvideo}, employs a dual-stream methodology MM-DiT that converts separate text and video streams into a flattened 3D space-time. 
\subsubsection{Single-Subject Personalization}
\label{sec:arch-single}
Single-subject personalization methods adapt a base one-shot video generator to faithfully reproduce a target individual’s appearance given only a one or a few reference frame as in \cite{liu2025phantom,polyak2024moviegen,gao2023dreamvideo,gu2023animatediff,yang2024cogvideox}. They achieve this by injecting or fine-tuning compact identity modules such as: (i) Embedding adapters inject lightweight projection modules into a frozen generator to encode subject appearance. Phantom~\cite{liu2025phantom} fine-tunes these adapters on a handful of exemplar frames to capture identity style, while MovieGen~\cite{polyak2024moviegen} applies per-subject adapters to preserve facial details across temporal generation. (ii) Textual inversion networks optimize new pseudo-token embeddings that encapsulate personalized identity concepts. DreamVideo~\cite{gao2023dreamvideo} learns these embeddings from static images to steer motion-conditioned synthesis, and CogVideoX~\cite{yang2024cogvideox} extends inversion across video frames for enhanced temporal consistency. (iii) LoRA layers ~\cite{hu2021lora} insert trainable low-rank matrices into attention modules, enabling efficient adaptation of large backbones. AnimateDiff~\cite{gu2023animatediff} leverages LoRA adapters in both U-Net and transformer blocks to personalize motion and appearance with minimal compute, avoiding full-model retraining.

\subsubsection{Multiple-Subject Personalization}
\label{sec:arch-one-multi}
Multi-subject personalization extends one-shot video synthesis to scenes with multiple distinct entities by integrating dedicated modules that encode and fuse each subject’s identity separately as in \cite{chen2025videoalchemist, chen2025skyreelsv2, huang2025conceptmaster, hu2025hunyuancustom}. Multi-subject personalization can be organized into four modular strategies: (i) Cross-Attention Fusion inserts per-entity attention heads into a frozen diffusion backbone so that each subject’s image and text descriptor attend separately. Video Alchemist binds each reference via dedicated cross-attention layers to support open-set multi-entity conditioning without fine-tuning~\cite{chen2025videoalchemist}. (ii) Element Embedding Fusion jointly encodes diverse scene elements into a unified latent space and injects them via fusion modules for coherent multi-entity synthesis. SkyReels-v2 learns an image–text joint embedding that precisely assembles multiple characters and backgrounds in one pass~\cite{chen2025skyreelsv2}. (iii) Decoupled Concept Embeddings learns independent latent vectors for each subject to avoid identity crosstalk, injecting them through diffusion-transformer adapters. ConceptMaster enforces strong per-entity disentanglement by adapting low-rank concept tokens~\cite{huang2025conceptmaster}. (iv) Multi-Modal Adapter Fusion layers modality-specific adapters (text, image, audio) into a unified fusion pipeline to preserve subject consistency across modalities. HunyuanCustom uses LLaVA-based text–image fusion and AudioNet adapters to maintain coherent identities in multi-subject, multi-modal videos~\cite{hu2025hunyuancustom}.

\subsubsection{Multi-Shot Narrative Planning}
\label{sec:arch-narrative}
Multi-shot narrative planning refers to the generation of long video clips explicitly segmented into discrete shots or scenes, ensuring coherent transitions and consistent visual elements across cuts as in \cite{gao2025seedance, chen2025skyreelsv2, huang2025stepvideo}. (i) End-to-end native planning as in Seedance~\cite{gao2025seedance}. This method integrates shot segmentation directly within its diffusion-transformer backbone via Multishot MM-RoPE and per-shot captions, producing all shots in a single pass with implicit cut boundaries. (ii) Planner-on-top architecture as in StepVideo~\cite{huang2025stepvideo}. This methodology adds a StoryAnchors layer on top of its Step-Video-T2V backbone; an LLM expands the prompt into a script, StoryAnchors predicts key “anchor” frames for each shot, and the base model animates between anchors to yield a coherent multi-shot sequence. (iii) LLM-directed autoregressive planning as in SkyReels-V2~\cite{chen2025skyreelsv2}. This architecture uses a multimodal language model to decompose the user prompt into a sequence of shot instructions, then a diffusion-forcing autoregressive generator renders shots one after another—supporting hard cuts, soft dissolves, and infinite-length film generation.


\subsection{Token-Stream Autoregressive-Token}
\label{sec:arch-token}

This method formulates video generation as next-token prediction over a unified text–video token stream. Specifically, a decoder-only transformer with causal attention autoregressively predicts each token conditioned on all previously generated tokens as in \cite{wang2024loong, singer2023videopoet}. VideoPoet~\cite{singer2023videopoet} adopts MagViT-v2~\cite{yu2023magvitv2} to tokenize video clips, whereas Loong \cite{wang2024loong} leverages a causal 3D-CNN encoder followed by vector quantization clustering to produce discrete video tokens. Both then apply a decoder-only transformer to predict the subsequent token in the combined sequence and finally employ super-resolution to recover spatial details. Despite their flexibility, these approaches incur high-frequency detail loss and compression artifacts in fine details. The need to attend over all prior tokens imposes substantial memory and compute overhead, leading to slow inference. Moreover, early tokens are harder to predict and error accumulates across which further degrades long-range temporal coherence and visual fidelity.


\subsection{Closed Source Video Generation}
\label{sec:arch-closed}
Proprietary video generation systems have pushed the boundaries of realism and complexity. Kling2.1~\cite{kling2p1}, Runway Gen-3~\cite{runwaygen3}, MiniMax Hailuo~\cite{hailuo2025}, Pika 2.2~\cite{pika1p5}, Sora~\cite{openai2024sora}, and MovieGen~\cite{polyak2024moviegen} have established a substantial performance gap between closed-source and open-source approaches. More recently, Google’s Veo3~\cite{deepmind2025veo3} and ByteDance’s Seedance 1.0~\cite{gao2025seedance} generate high-fidelity videos that adhere to physical laws, support complex multi-subject scenes, capture intricate pose dynamics and cinematic camera movements, and maintain character consistency. 

\setlength{\tabcolsep}{0pt}

\definecolor{whiteshade}{gray}{1}
\definecolor{rowshade}{gray}{0.90}
\definecolor{headershade}{gray}{0.95}

\newcommand{\applyshade}[1]{\cellcolor{rowshade}#1}
\newcommand{\applywhite}[1]{\cellcolor{white}#1}

\newcommand{\spacercolshade}{\multicolumn{1}{p{0.15cm}}{\cellcolor{rowshade}}}
\newcommand{\spacercolwhite}{\multicolumn{1}{p{0.15cm}}{\cellcolor{white}}}

\newcolumntype{Y}{>{\RaggedRight\arraybackslash}p{0.1275\textwidth}}

\begin{strip}
  \centering
  \scriptsize
  \captionof{table}{
    A comprehensive backbone table of recent video generation frameworks, organized by their core training objectives, that reveals emerging design patterns and architectural innovations across state-of-the-art systems.
  }
  \label{tab:backbone1}
  \renewcommand{\arraystretch}{1.3} 

    \begin{tabular}{ 
        p{0.1\textwidth} 
        Y                
        Y                
        Y                
        Y                
        Y                
        Y                
        Y                
      }
    \rowcolor{headershade}
    \cellcolor{white}{\small\bfseries Training Objective} 
    & \multicolumn{1}{Y}{\small\bfseries Paper}
    & \multicolumn{1}{Y}{\small\bfseries\makecell[l]{Backbone}}
    & \multicolumn{1}{Y}{\small\bfseries\makecell[l]{Text-Visual \\Tower}}
    & \multicolumn{1}{Y}{\small\bfseries\makecell[l]{Visual-Video \\Tower}}
    & \multicolumn{1}{Y}{\small\bfseries\makecell[l]{\hspace{0.5cm}Positional\\\hspace{0.5cm}Encodings}}
    & \multicolumn{1}{Y}{\small\bfseries\makecell[l]{\hspace{0.5cm}Params}}
    & \multicolumn{1}{Y}{\small\bfseries Resolution} \\
    \toprule


    \rowcolor{rowshade}
    \cellcolor{white}{} 
    & \applyshade{Seedance~\cite{gao2025seedance}}
    & \applyshade{MM-DiT}
    & \applyshade{\makecell[l]{Qwen2.5-14B}}
    & \applyshade{\makecell[l]{VAE}}
    & \applyshade{\makecell[l]{\hspace{0.5cm}3DRoPE\\ \hspace{0.5cm}MM-RoPE}}
    & \applyshade{\makecell[l]{\hspace{0.5cm}--}}
    & \applyshade{720p, 1080p} \\

    {} 
    & \applywhite{HunyuanVideo-Avatar~\cite{chen2025hunyuanvideoavatar}}
    & \applywhite{MM-DiT}
    & \applywhite{\makecell[l]{LLaVA}}
    & \applywhite{\makecell[l]{Two Hunyuan \\3D VAE}}
    & \applywhite{\makecell[l]{\hspace{0.5cm}3DRoPE}}
    & \applywhite{\makecell[l]{\hspace{0.5cm}13B}}
    & \applywhite{704p, 1216p} \\

    \rowcolor{rowshade}
    \cellcolor{white}{}
    & \applyshade{MAGI-1~\cite{teng2025magi1}}
    & \applyshade{DiT}
    & \applyshade{\makecell[l]{T5}} 
    & \applyshade{\makecell[l]{Transformer-based\\VAE}} 
    & \applyshade{\makecell[l]{\hspace{0.5cm}3DRoPE}}
    & \applyshade{\makecell[l]{\hspace{0.5cm}4.5B--24B}}
    & \applyshade{720p} \\

    {}
    & \applywhite{HunyuanCustom~\cite{hu2025hunyuancustom}}
    & \applywhite{Hunyuan-MM-DiT}
    & \applywhite{\makecell[l]{LLaVA}} 
    & \applywhite{\makecell[l]{Two Hunyuan\\3D VAE}} 
    & \applywhite{\makecell[l]{\hspace{0.5cm}3DRoPE}} 
    & \applywhite{\makecell[l]{\hspace{0.5cm}13B}}
    & \applywhite{512p, 720p} \\

    \rowcolor{rowshade}
    \cellcolor{white}{}
    & \applyshade{Veo3~\cite{deepmind2025veo3}}
    & \applyshade{DiT}
    & \applyshade{--}
    & \applyshade{--} 
    & \applyshade{\hspace{0.5cm}--} 
    & \applyshade{\makecell[l]{\hspace{0.5cm}--}}
    & \applyshade{1080p} \\

    {}
    & \applywhite{SkyReels-v2~\cite{chen2025skyreelsv2}}
    & \applywhite{Wan-DiT}
    & \applywhite{\makecell[l]{umT5}} 
    & \applywhite{\makecell[l]{Wan VAE}}
    & \applywhite{\makecell[l]{\hspace{0.5cm}Learnable\\\hspace{0.5cm}Frequency\\ \hspace{0.5cm}Embeddings}} 
    & \applywhite{\makecell[l]{\hspace{0.5cm}1.3B, 5B, 14B}}
    & \applywhite{\makecell[l]{256p, 360p,\\540p, 720p}} \\

    \rowcolor{rowshade}
    \cellcolor{white}{}
    & \applyshade{Open-Sora 2.0~\cite{peng2025opensora2}}
    & \applyshade{Flux (MM-DiT)}
    & \applyshade{\makecell[l]{T5-XXL,\\CLIP-Large}} 
    & \applyshade{\makecell[l]{HunyuanVideo\\3DVAE,\\Video Deep\\Compression\\Autoencoder}} 
    & \applyshade{\makecell[l]{\hspace{0.5cm}3DRoPE}} 
    & \applyshade{\makecell[l]{\hspace{0.5cm}11B}}
    & \applyshade{256p, 768p} \\

    {}
    & \applywhite{WAN2.1\cite{wan2025}}
    & \applywhite{DiT + Cross-attn}
    & \applywhite{\makecell[l]{umT5,\\Qwen2-VL}} 
    & \applywhite{\makecell[l]{Wan-VAE}}
    & \applywhite{\makecell[l]{\hspace{0.5cm}Standard\\\hspace{0.5cm}Sinusoidal\\ \hspace{0.5cm}Spatial positional\\ \hspace{0.5cm}Encodings}} 
    & \applywhite{\makecell[l]{\hspace{0.5cm}1.3B, 14B}}
    & \applywhite{480p, 720p} \\

    \cellcolor{white}\multirowcell{2}[0pt][l]{\textbf{Flow-}\\\textbf{Matching}} 
    & \applyshade{VACE~\cite{vace2025}}
     & \applyshade{Wan-T2V-14B, LTX-Video-2B}
    & \applyshade{Inherited}
    & \applyshade{Inherited}
    & \applyshade{\makecell[l]{\hspace{0.5cm}Inherited}}
    & \applyshade{\makecell[l]{\hspace{0.5cm}2B, 14B}}
    & \applyshade{480p, 720p}      \\

    & \applywhite{Phantom~\cite{liu2025phantom}}
    & \applywhite{MMDiT}
    & \applywhite{\makecell[l]{T5\\Dinov2 (Ref.~Img)}}
    & \applywhite{\makecell[l]{(CLIP, VAE)\\(Qwen2.5, 3DVAE)}}
    & \applywhite{\makecell[l]{\hspace{0.5cm}3DRoPE}}
    & \applywhite{\makecell[l]{\hspace{0.5cm}1.3B, 14B}}
    & \applywhite{480p, 720p}      \\

    \rowcolor{rowshade}
    \cellcolor{white}{}
    & \applyshade{StepVideo~\cite{huang2025stepvideo}}
    & \applyshade{DiT}
    & \applyshade{\makecell[l]{Hunyuan-CLIP,\\Step-LLM}}
    & \applyshade{\makecell[l]{Video-VAE}} 
    & \applyshade{\makecell[l]{\hspace{0.5cm}3DRoPE}} 
    & \applyshade{\makecell[l]{\hspace{0.5cm}30B}}
    & \applyshade{544p} \\

    {}
    & \applywhite{ConceptMaster~\cite{huang2025conceptmaster}}
    & \applywhite{\makecell[l]{Transformer-based\\latent diffusion}}
    & \applywhite{\makecell[l]{T5\\CLIP}}
    & \applywhite{\makecell[l]{3DVAE}} 
    & \applywhite{\makecell[l]{\hspace{0.5cm}3D self-attention}} 
    & \applywhite{\makecell[l]{\hspace{0.5cm}--}}
    & \applywhite{--} \\

    \rowcolor{rowshade}
    \cellcolor{white}{}
    & \applyshade{VideoAlchemist~\cite{chen2025videoalchemist}}
    & \applyshade{DiT}
    & \applyshade{\makecell[l]{DiT Text Encoder,\\CLIP,\\Arcface}}
    & \applyshade{\makecell[l]{(CogVideoX-5B VAE,\\DiT Tokenizer),\\(CLIP ViT-L/14,\\DINOv2 ViT-L/14)}} 
    & \applyshade{\makecell[l]{\hspace{0.5cm}RoPE}}
    & \applyshade{\makecell[l]{\hspace{0.5cm}5B}}
    & \applyshade{256p} \\

    {}
    & \applywhite{HunyuanVideo~\cite{kong2024hunyuanvideo}}
    & \applywhite{Flux (MM-DiT)}
    & \applywhite{\makecell[l]{Hunyuan MLLM\\Decoder,\\CLIP}} 
    & \applywhite{\makecell[l]{3D VAE}} 
    & \applywhite{\makecell[l]{\hspace{0.5cm}3DRoPE}} 
    & \applywhite{\makecell[l]{\hspace{0.5cm}13B}}
    & \applywhite{720p} \\

    \rowcolor{rowshade}
    \cellcolor{white}{}
    & \applyshade{LTX-video~\cite{hacohen2025ltxvideo}}
    & \applyshade{DiT + Cross-attn}
    & \applyshade{\makecell[l]{DiT Text Encoder}} 
    & \applyshade{\makecell[l]{Video-VAE}} 
    & \applyshade{\makecell[l]{\hspace{0.5cm}RoPE}} 
    & \applyshade{\makecell[l]{\hspace{0.5cm}2B}}
    & \applyshade{512p} \\

    {}
    & \applywhite{MovieGen~\cite{polyak2024moviegen}}
    & \applywhite{LLaMa3 Design}
    & \applywhite{\makecell[l]{UL2,\\ByT5,\\Long-prompt\\MetaCLIP}}
    & \applywhite{\makecell[l]{TAE,\\VAE (Spatial\\Upsampler)}} 
    & \applywhite{\makecell[l]{\hspace{0.5cm}Factorized}} 
    & \applywhite{\makecell[l]{\hspace{0.5cm}30B}}
    & \applywhite{256p, 1080p} \\

    \rowcolor{rowshade}
    \cellcolor{white}{}
    & \applyshade{Pyramid Flow~\cite{jin2024pyramidflow}}
    & \applyshade{MM-DiT}
    & \applyshade{--} 
    & \applyshade{\makecell[l]{Pyramid Stages\\Autoregressive\\Temporal Pyramid}} 
    & \applyshade{\hspace{0.5cm}--} 
    & \applyshade{\makecell[l]{\hspace{0.5cm}--}}
    & \applyshade{768p} \\

    {}
    & \applywhite{Sora~\cite{openai2024sora}}
    & \applywhite{DiT}
    & \applywhite{--} 
    & \applywhite{--} 
    & \applywhite{\hspace{0.5cm}--} 
    & \applywhite{\makecell[l]{\hspace{0.5cm}--}}
    & \applywhite{480p, 1080p} \\
    \hline

    \cellcolor{white}\multirowcell{2}[0pt][l]{\\\textbf{Score-}\\\textbf{Matching}} 
    & \applyshade{SVD~\cite{ho2023svd}}
    & \applyshade{3DUNet}
    & \applyshade{CLIP}
    & \applyshade{SD 2D VAE}
    & \applyshade{\makecell[l]{\hspace{0.5cm}--}}
    & \applyshade{\makecell[l]{\hspace{0.5cm}1.5B}}
    & \applyshade{512p} \\

    
    & \applywhite{I2VGen-XL~\cite{zhang2023i2vgenxl}}
    & \applywhite{3DUNet}
    & \applywhite{CLIP}
    & \applywhite{VQGAN}
    & \applywhite{\makecell[l]{\hspace{0.5cm}Standard VLDM\\\hspace{0.5cm}positional\\\hspace{0.5cm}embeddings}}
    & \applywhite{\makecell[l]{\hspace{0.5cm}--}}
    & \applywhite{64p, 720p} \\
    \hline


    & \applyshade{StoryDiffusion~\cite{zhou2024storydiffusion}}
    & \applyshade{UNet}
    & \applyshade{CLIP}
    & \applyshade{SD 2D VAE}
    & \applyshade{\makecell[l]{\hspace{0.5cm}--}}
    & \applyshade{\makecell[l]{\hspace{0.5cm}1B, 4B}}
    & \applyshade{512p} \\
    

    \cellcolor{white}\multirowcell{2}[0pt][l]{\textbf{}\\\\\textbf{DDIM}} 
    & \applywhite{DreamVideo~\cite{gao2023dreamvideo}}
    & \applywhite{3DUNet}
    & \applywhite{CLIP}
    & \applywhite{LDM VAE}
    & \applywhite{\makecell[l]{\hspace{0.5cm}Standard\\\hspace{0.5cm}Transformer\\\hspace{0.5cm}positional embed}}
    & \applywhite{\makecell[l]{\hspace{0.5cm}85M}}
    & \applywhite{--} \\
    

    & \applyshade{Ca2-VDM~\cite{gao2024ca2vdm}}
    & \applyshade{UNet}
    & \applyshade{T5}
    & \applyshade{SD 2D VAE}
    & \applyshade{\makecell[l]{\hspace{0.5cm}SinusoidalSpatial\\\hspace{0.5cm}Positional Embed,\\ \hspace{0.5cm}Temporal\\\hspace{0.5cm}Positional Embed\\\hspace{0.5cm}with cyclic-shift \\\hspace{0.5cm}mechanism}}
    & \applyshade{\makecell[l]{\hspace{0.5cm}--}}
    & \applyshade{256p} \\
    \hline

  \end{tabular}
  \renewcommand{\arraystretch}{1} 
\end{strip}


\addtocounter{table}{-1}
\begin{strip}
  \centering
  \scriptsize
  \captionof{table}{
    Continued
  }
  \label{tab:backbone1}
  \renewcommand{\arraystretch}{1.3} 

    \begin{tabular}{ 
        p{0.1\textwidth} 
        Y                
        Y                
        Y                
        Y                
        Y                
        Y                
        Y                
      }
    \rowcolor{headershade}
    \cellcolor{white}{\small\bfseries Training Objective} 
    & \multicolumn{1}{Y}{\small\bfseries Paper}
    & \multicolumn{1}{Y}{\small\bfseries\makecell[l]{Backbone}}
    & \multicolumn{1}{Y}{\small\bfseries\makecell[l]{Text-Visual \\Tower}}
    & \multicolumn{1}{Y}{\small\bfseries\makecell[l]{Visual-Video \\Tower}}
    & \multicolumn{1}{Y}{\small\bfseries\makecell[l]{\hspace{0.5cm}Positional\\\hspace{0.5cm}Encodings}}
    & \multicolumn{1}{Y}{\small\bfseries\makecell[l]{\hspace{0.5cm}Params}}
    & \multicolumn{1}{Y}{\small\bfseries Resolution} \\
    \toprule

    & \applyshade{Lumiere~\cite{bartal2024lumiere}}
    & \applyshade{Space-Time U-Net}
    & \applyshade{Imagen T5-XXL}
    & \applyshade{Pixel space}
    & \applyshade{\makecell[l]{\hspace{0.5cm}--}}
    & \applyshade{\makecell[l]{\hspace{0.5cm}--}}
    & \applyshade{--} \\
    

    \cellcolor{white}\multirowcell{2}[0pt][l]{\\\\\textbf{DDPM}}
    & \applywhite{SEINE~\cite{chen2023seine}}
    & \applywhite{LaVie-UNet}
    & \applywhite{SD}
    & \applywhite{SD VQGAN/VQVAE}
    & \applywhite{\makecell[l]{\hspace{0.5cm}--}}
    & \applywhite{\makecell[l]{\hspace{0.5cm}--}}
    & \applywhite{320p} \\


    & \applyshade{AnimateDiff~\cite{gu2023animatediff}}
    & \applyshade{3DUNet}
    & \applyshade{CLIP,Conditioned Cross-Attention}
    & \applyshade{SD VAE}
    & \applyshade{\makecell[l]{\hspace{0.5cm}Standard\\\hspace{0.5cm}Transformer\\ \hspace{0.5cm}Temporal\\\hspace{0.5cm}Positional Embed}}
    & \applyshade{\makecell[l]{\hspace{0.5cm}--}}
    & \applyshade{512p} \\


    & \applywhite{VDM~\cite{ho2022vdm}}
    & \applywhite{3DUNet}
    & \applywhite{--}
    & \applywhite{Pixel space}
    & \applywhite{\makecell[l]{\hspace{0.5cm}Relative Position\\\hspace{0.5cm}Embed}}
    & \applywhite{\makecell[l]{\hspace{0.5cm}--}}
    & \applywhite{64p, 128p} \\
    \hline

    \rowcolor{white}
    \cellcolor{white}\multirowcell{2}[0pt][l]{\textbf{V-Prediction}\\\textbf{\& Zero-SNR}}
    & \applyshade{CogVideoX~\cite{yang2024cogvideox}}
    & \applyshade{DiT + Cross-attn}
    & \applyshade{T5}
    & \applyshade{3DVAE}
    & \applyshade{\makecell[l]{\hspace{0.5cm}3DRoPE}}
    & \applyshade{\makecell[l]{\hspace{0.5cm}2B, 5B}}
    & \applyshade{768p} \\
    
    
    & \applyshade{}
    & \applyshade{}
    & \applyshade{}
    & \applyshade{}
    & \applyshade{}
    & \applyshade{}
    & \applyshade{} \\
    \hline

    \cellcolor{white}\multirowcell{2}[0pt][l]{\textbf{Reconstruction}\\\textbf{Loss}}
    & \applywhite{Open-Sora Plan~\cite{lin2024opensoraplan}}
    & \applywhite{UNet Skiparse Denoiser}
    & \applywhite{mT5-XXL}
    & \applywhite{Wavelet-Flow VAE}
    & \applywhite{\makecell[l]{\hspace{0.5cm}3DRoPE}}
    & \applywhite{\makecell[l]{\hspace{0.5cm}--}}
    & \applywhite{256p} \\
    \hline

    \cellcolor{white}\multirowcell{2}[0pt][l]{\textbf{Next-Token}\\\textbf{Prediction}}
    & \applyshade{Loong~\cite{wang2024loong}}
    & \applyshade{LLaM decoder design}
    & \applyshade{--} 
    & \applyshade{\makecell[l]{3DCNN + Clustering\\Vector Quantization}}
    & \applyshade{\makecell[l]{\hspace{0.5cm}Causal Unidirect.\\\hspace{0.5cm}Attention Across\\ \hspace{0.5cm}Token Sequence}}
    & \applyshade{\makecell[l]{\hspace{0.5cm}700M, 3B, 7B}}
    & \applyshade{--} \\ 
    \hline

    \cellcolor{white}\multirowcell{2}[0pt][l]{\textbf{Autoregressive}\\\textbf{Multimodal}\\\textbf{Tokenization}}
    & \applywhite{VideoPoet~\cite{singer2023videopoet}}
    & \applywhite{Decoder-only Transformer LLM}
    & \applywhite{T5}
    & \applywhite{MAGVIT-V2}
    & \applywhite{\makecell[l]{\hspace{0.5cm}Standard\\\hspace{0.5cm}Transformer\\\hspace{0.5cm}Positonal Embed}}
    & \applywhite{\makecell[l]{\hspace{0.5cm}300M, 8B}}
    & \applywhite{128p} \\

    
    & \applywhite{}
    & \applywhite{}
    & \applywhite{}
    & \applywhite{}
    & \applywhite{}
    & \applywhite{}
    & \applywhite{} \\
    \hline

    \cellcolor{white}\multirowcell{2}[0pt][l]{\textbf{Masked}\\\textbf{Token}}
    & \applyshade{MAGVIT-v2~\cite{yu2023magvitv2}}
    & \applyshade{MLLM}
    & \applyshade{--} 
    & \applyshade{3DCNN-VQVAE}
    & \applyshade{--} 
    & \applyshade{\makecell[l]{\hspace{0.5cm}300M}}
    & \applyshade{256p, 512p} \\


    & \applyshade{}
    & \applyshade{}
    & \applyshade{}
    & \applyshade{}
    & \applyshade{}
    & \applyshade{}
    & \applyshade{} \\

    \hline 
    
  \end{tabular}
  \renewcommand{\arraystretch}{1}
\end{strip}
\section{Long Video Generation Architectural Modules Recommendations}
\label{sec:architecture}

In this section, we illustrate the important components of the video generation architecture and we recommend the usage of these components. For example, we discuss the usage of text-encoder in literature and recommend using MLLM, while recent research used variations of T5 alongside CLIP. We also recommend using MeanFlow as a training objective for the diffusion model. While for the architectural main backbone, we recommend using MM-DiT and Flux-MM-DiT. Even though we recommend specific components, the architecture can be different based on the taxonomy discussed in Figure ~\ref{fig:main-taxonomy}. A summary of this survey is shown in Table ~\ref{tab:backbone1} comparing the key architectural components used by each literature.

\subsection{Text-Visual Encoder}
Text-visual encoder is used to extract the text embeddings and extract the similarity score between text and image embeddings. It is common between the literature to use CLIP’s text-visual encoder ~\cite{clip2021} in conjunction with T5, T5-XXL or umT5 ~\cite{t5-2019, umt5-2023, t5-xxl-2024} as they provided robust text–to-video alignment by extracting semantically rich embeddings. Recently, HunyuanVideo ~\cite{kong2024hunyuanvideo} replaced T5 with a Multimodal Large Language Model (MLLM) reaching better alignment between visual features and text embeddings. The equation of the clip encoder is as follows:
\begin{align}
h_I &= \mathrm{ViT}_I\!\bigl(\mathrm{PatchEmbed}(I)\bigr), \\
h_T &= \mathrm{Tr}_T\!\bigl(\mathrm{TokEmbed}(T)\bigr), \\
\mathcal{L}_{\text{clip}} &=
-\log\frac{\exp\bigl(\tau^{-1}h_I^{\!\top}h_T\bigr)}
     {\sum_{j}\exp\bigl(\tau^{-1}h_I^{\!\top}h_{T}^{(j)}\bigr)} .
\end{align}
Where $I$ RGB image, $T$ caption text, $h_I,h_T$ CLS embeddings, $\tau$ temperature scalar, $\mathcal{L}_{\text{clip}}$ contrastive loss. 
While for T5 equations:
\begin{align}
H_E &= \mathrm{T5\text{-}Enc}\!\bigl(E_{\text{tok}}(T_{\text{src}})\bigr), \\
h_{D,i} &= \mathrm{T5\text{-}Dec}\!\bigl(E_{\text{tok}}(Y_{<i}),H_E\bigr), \\
\mathcal{L}_{\text{T5}} &=
-\sum_{i}\log\!\left[\operatorname{softmax}(W_O h_{D,i})_{\,y_i}\right].
\end{align}
Where $T_{\text{src}}$ input tokens, $Y$ target tokens with prefix $Y_{<i}$, $E_{\text{tok}}$ token–embedding table, $H_E$ encoder context, $h_{D,i}$ decoder hidden state at step $i$, $W_O$ output projection, $\mathcal{L}_{\text{T5}}$ autoregressive cross-entropy loss.

\subsection{Diffusion Training Objective} 
Diffusion models have demonstrated remarkable success in high-quality image synthesis, laying the groundwork for video generation by leveraging denoising diffusion models such as DDPM ~\cite{ddpm2020} and DDIM ~\cite{ddim2020}. The introduction of a latent representation via autoencoders VAE further reduced computational cost and enabled more efficient sampling as in ~\cite{ldm2022}. Flow matching methods have emerged as a more robust and efficient alternative to denoising diffusion samplers like DDPM and DDIM by directly regressing continuous-time vector fields that transport noise to data, thus reducing reliance on multi-step noise schedules ~\cite{flowmatching2023, flowmatching2024}. More recently, MeanFlow~\cite{geng2025meanflows} replaces instantaneous ordinary differential equation (ODE) velocities with a learned average velocity field. On Kinetics-400, it achieves an FVD of 128, compared to 142 for flow-matching. On UCF-101, it attains an SSIM of 0.85~\cite{wang2004image}, exceeding flow-matching’s 0.82~\cite{davtyan2023river}. It also achieves an LPIPS of 0.12~\cite{zhang2018unreasonable}, improving on flow-matching’s 0.15~\cite{flowmatching2023, flowmatching2024}. MeanFlow reduces inference time by 4× compared to flow-matching. It also narrows the quality gap with multi-step diffusion models~\cite{wang2025diffuse}. Based on these results, we recommend using MeanFlow for efficient, high-quality video generation.

Flow-matching formulas are as follows:
\begin{align}
\mathbf{u}_{t\!\rightarrow\!t+1} &= 
\mathcal{F}\!\bigl(h_I^{(t)},h_I^{(t+1)};\theta_{\mathcal{F}}\bigr), \\
\mathcal{L}_{\text{flow}} &=
\bigl\lVert\mathbf{u}_{t\!\rightarrow\!t+1}(p)-\hat{\mathbf{u}}_{t\!\rightarrow\!t+1}(p)\bigr\rVert_{2}^{2}.
\end{align}
Where $h_I^{(t)}$ image feature at frame $t$, $\mathbf{u}_{t\to t+1}$ predicted optical flow, $\hat{\mathbf{u}}_{t\to t+1}$ ground-truth flow, $\mathcal{F}$ flow network with parameters $\theta_{\mathcal{F}}$, $\mathcal{L}_{\text{flow}}$ L1 loss. While for MeanFlow formula:
\begin{align}
\bar{\mathbf{u}} &=
\frac{1}{(T-1)\,H\,W}\sum_{t=1}^{T-1}\sum_{p=1}^{HW}
\mathbf{u}_{t\!\rightarrow\!t+1}(p).
\end{align}
Where $T$ video length in frames, $H,W$ frame height and width, $p$ linear pixel index, $\mathbf{u}_{t\to t+1}(p)$ optical-flow vector at pixel $p$ between frames $t$ and $t\!+\!1$, $\bar{\mathbf{u}}$ spatial–temporal average flow. 

\subsection{Variational Auto Encoder (VAE)}
VAE is crucial in learning the compressed latent format of original visual data. In video generation, the 3D VAE captures complex spatio-temporal dependencies. Some designs choose to replace standard VAE with VQ-VAE ~\cite{VQVAE2017, VQVAE2019} to enhance the compression and construction. Other designs ~\cite{hacohen2025ltxvideo} modified the VAE decoder to fit-in a last denoising step while converting the latents to pixels. 3D VAE is the most common among top-performing literature.
\begin{align}
z &\sim q_\phi(z\!\mid\!X)= \mathcal{N}\!\bigl(\mu_\phi(X),\operatorname{diag}\sigma_\phi^{2}(X)\bigr), \\
\mathcal{L}_{\text{3DVAE}} &=
\mathbb{E}_{q_\phi}\!\bigl[\lVert D_\psi(z)-X\rVert_2^{2}\bigr]+
\beta\,D_{\mathrm{KL}}\!\bigl(q_\phi\|p(z)\bigr).
\end{align}
Where $X$ ground-truth video tensor, $q_\phi$ probabilistic encoder, $\mu_\phi,\sigma_\phi$ per-frame Gaussian parameters, $z$ latent code, $D_\psi$ 3D decoder, $p(z)$ standard normal prior, $\beta$ KL-weight, $\mathcal{L}_{\text{3DVAE}}$ reconstruction + regularization loss.

\subsection{Dual Variational Auto Encoder}
Recent architectures decouple appearance and motion by using two distinct encoders. One for static image features and another for temporal dynamics like in ~\cite{chen2025hunyuanvideoavatar, hu2025hunyuancustom, peng2025opensora2, liu2025phantom, chen2025videoalchemist, polyak2024moviegen}. This enables specialized feature learning and reducing interference between modalities. Models like Open-Sora 2.0 ~\cite{peng2025opensora2} adopt this dual-stream design to lower training costs by 5–10× while maintaining state-of-the-art video quality. Similarly, VideoAlchemist ~\cite{chen2025videoalchemist} demonstrates built-in multi-subject personalization and improved identity consistency without test-time fine-tuning by leveraging separate foreground and background encoding streams.

\subsection{Attention Mechanism}
Video Diffusion Models (VDM) extended the 2D UNet architecture ~\cite{unet2015} into 3D UNet to model spatio-temporal (2D spatial + 1D temporal layer) dependencies directly~\cite{ho2022vdm}. Subsequent methods such as AnimatedDiff~\cite{gu2023animatediff}, MagicVideo~\cite{magicvideo2022}, ModelScope Text-to-Video~\cite{modelscope2023}, Stable Video Diffusion (SVD)~\cite{ho2023svd}, and CogVideoX~\cite{yang2024cogvideox} adopted a hybrid strategy by integrating 1D temporal attention blocks into a 2D spatial backbone of the attention, yielding full 3D attention for improved frame coherence. Despite these advances, early diffusion-based video systems typically generated  2–5 second clips with artifacts and limited long-term consistency, indicating that simple 2D + 1D temporal modules remain insufficient for robust motion modeling. 

Recent models with robust performance as in Seedance 1.0 ~\cite{gao2025seedance}, they decoupled the attention spatial and temporal layers, where spatial layers perform global self-attention within each frame and temporal layers apply window-partitioned 3D self-attention across time to link frames causally. Seedance incorporates specialized windowed attention modules in temporal layers, partitioning frames into local blocks that attend within sliding windows, yielding 10× faster inference on 1080p benchmarks. Another top performing method by MAGI-1 ~\cite{teng2025magi1}, employs block-causal self-attention, which performs unrestricted full attention within each fixed-length video chunk while applying causal masks across chunk boundaries. MAGI-1 integrates a Flexible-Flash-Attention kernel on top of FlashAttention-3 ~\cite{shah2024flashattention3}, optimizing memory access patterns and reducing GPU communication overhead during attention computation. MAGI-1 further accelerates computation by computing shared query projections once and feeding them into spatial-temporal self-attention and cross-attention blocks in parallel.
\subsection{Positional Encoding}
WAN2.1~\cite{wan2025} retains standard sinusoidal encodings to reduce computational overhead while LTX-Video~\cite{hacohen2025ltxvideo} applies one-dimensional rotary positional embeddings (RoPE) over flattened tokens for real-time inference. Recently, Three-dimensional rotary positional embeddings (3D RoPE) that rotate feature pairs across time and space have been used by HunyuanVideo~\cite{kong2024hunyuanvideo}, MAGI-1~\cite{teng2025magi1}, StepVideo~\cite{huang2025stepvideo}, HunyuanCustom~\cite{hu2025hunyuancustom}, Phantom~\cite{liu2025phantom} and Open-Sora~2.0~\cite{peng2025opensora2} to enhance motion coherence and length extrapolation. Seedance~\cite{gao2025seedance} introduces a multi-modal RoPE (MM-RoPE) by appending it to oridnary 3D RoPE for caption tokens, which tightens text–video alignment in multi-shot generation which proves to be a promising technique.
\subsection{Diffusion-based Transformers}
The transformer-based backbones such as Diffusion Transformers (DiT) ~\cite{dit2023}, Latte ~\cite{latte2025}, PixArt-$\alpha$ ~\cite{pixart2023} are able to generate much better images and videos than UNet backbones. DiT operating on latent image patches, leveraging global cross-attention for conditioning and yielding to high-fidelity videos as in MAGI-1~\cite{teng2025magi1}, StepVideo~\cite{huang2025stepvideo}, Wan2.1~\cite{wan2025}, LTX-Video ~\cite{hacohen2025ltxvideo} and VideoAlchemist~\cite{chen2025videoalchemist}. That is proceeded with MultimModal Diffusion Transformer (MM-DiT) ~\cite{flowmatching2024}, a dual-stream DiT architecture, as the text embeddings are concatenated with the visual embeddings to have a linked text-visual attention which yields stronger text–video alignment and lower FID at the cost of increased parameter count and inference overhead as in Seedance 1.0 ~\cite{gao2025seedance}, Phantom ~\cite{liu2025phantom} and PyramidFlow ~\cite{jin2024pyramidflow}. Another recent method is Flux-MM-DiT ~\cite{fluxdit2024}, which augments MM-DiT with rectified flow residual modules to enable one-step sampling and faster convergence, achieving comparable sample quality in far fewer denoising steps while introducing additional architectural complexity as in HunyuanVideo~\cite{kong2024hunyuanvideo} and Open-Sora 2.0~\cite{peng2025opensora2}.

\subsection{Prompt Enhancement}
User‐supplied prompts are often short, whereas training captions are multi‐sentence and richly detailed, causing a distribution mismatch that degrades video quality. To bridge this gap, an LLM‐powered prompt rewriting stage is added into the text–visual tower. For example, Seedance leverages Qwen2.5-Plus to expand concise user inputs with spatial, lighting, and action modifiers~\cite{gao2025seedance}; HunyuanVideo-Avatar uses LLaVA to paraphrase free-form queries into training‐style captions, reducing semantic drift~\cite{chen2025hunyuanvideoavatar}; and StepVideo incorporates a bespoke Step-LLM module to structure prompts into chunk‐level directives, ensuring smoother motion and coherent long‐form generation~\cite{huang2025stepvideo}. Models without dedicated rewrite modules such as VACE~\cite{vace2025} and VideoAlchemist~\cite{chen2025videoalchemist}, rely solely on fixed text encoders. By aligning rewritten prompts with the training caption distribution, these enhancements sharpen visual detail, suppress flicker, and unify style across video frames.

\subsection{Story Agent}
Operating at the narrative level, the story agent uses LLM to segment the input plot into scenes and shots, aligns characters, locations, and camera cues across those shots, and generates scene-specific prompts that preserve temporal coherence and multi-subject consistency. Following the approach of StoryDiffusion~\cite{zhou2024storydiffusion}, MovieAgent~\cite{movieagent2025}, and AutoStory~\cite{liu2023autostory}, these prompts drive an image-to-video stage for each keyframe; the resulting clips are concatenated into a seamless long-form video, yielding a decoupled architecture in which prompt-level refinements boost frame quality while the agent governs plot flow.
\section{Conclusions and Future Work}
Despite a narrowing performance gap between proprietary and open‐source video generation systems, closed‐source solutions still lead in overall quality. Recent open‐source models such as HunyuanVideo~\cite{kong2024hunyuanvideo} and Wan2.1~\cite{wan2025} demonstrate that open frameworks can now generate realistic, high‐fidelity videos.

\textbf{Our architectural analysis reveals that:} (i) MM‐DiT and Flux‐MM‐DiT serve as the most effective backbones for modern video diffusion (ii) Flow matching has supplanted DDIM and DDPM as the preferred diffusion training objective for realism (iii) MeanFlow generates promising results that may replace Flow-matching (iv) MLLMs outperform T5 as text encoders  (v) convolutional VAEs with discriminator loss remain superior for image and video encoding (vi) Dual VAE for image and video separately showed superior results compared to one VAE for both (vii) Dual usage of 3DRoPE and 3D MM‐RoPE as positional encodings yields better temporal coherence than traditional sinusoidal embeddings (viii) LLM‐driven prompt rewriting consistently enhances generation quality.

\textbf{The current limitations are observed as follows:} (i) Substantial memory and GPU requirements that limit model scale and clip length (ii) A shortage of open‐source long‐form video datasets suited to foundational generation tasks (iii) Existing annotated datasets lack critical metadata, such as camera shot styles, camera movements, and interpersonal relationships (v) Inconsistent temporal coherence, where frame‐to‐frame continuity breaks down in longer sequences (vi) Lack of fine‐grained control over semantics in object interactions beyond coarse prompts (vii) Difficulty in modeling multiple subjects simultaneously with reference images, resulting in identity inconsistencies and unrealistic interactions (viii) Generated videos are very short between 5-16 seconds (ix) Lack of story coherent generated videos

\textbf{Solutions for the creation of long-video generation and future work:} (i) Collect long-video open-source dataset (ii) Define and annotate a hierarchical metadata schema with four key pillars: narrative segments, cinematic shot labels, character attributes (pose \& emotion), and interaction graphs (iii) Quantization and pruning for the models to overcome resources limitations (iv) Models distillation to learn from teacher models (v) Integration of prompt enhancer (v) Dividing the prompt into story narration for the coherence of the video (vi) Using multiple adapters for personalization consistency (vii) Repeating the reference image across the spatiotemporal attention.

These insights collectively chart the progress in video generation and highlight key directions for future research aimed at bridging the remaining gaps.

\label{sec:conc}


{\small
    \bibliographystyle{ieeenat_fullname}
    \bibliography{main}
}

\clearpage
\appendix

\section*{Appendix}
\label{sec:appendix}

\setcounter{figure}{1} 
\begin{figure*}[t]
  \centering
  \includegraphics[width=\textwidth]{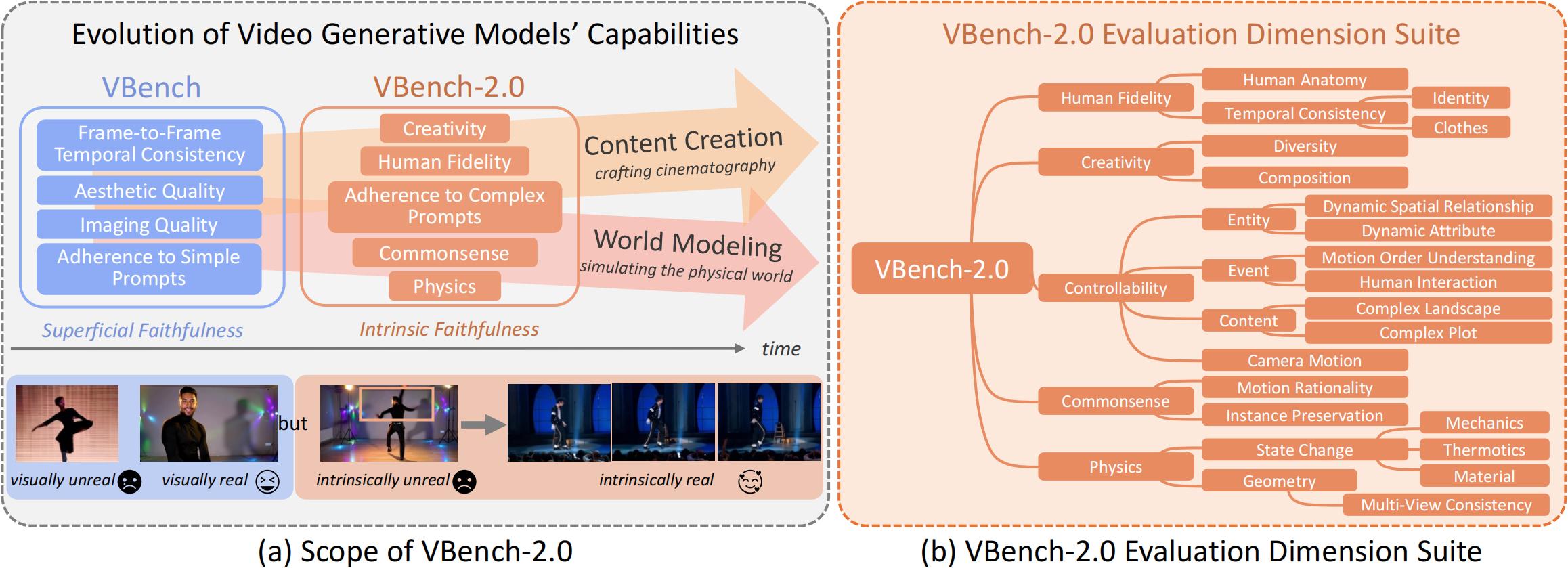}
  \caption{Overview of VBench evaluation metrics \cite{Huang2024}. VBench measures visual quality, motion smoothness, identity consistency, temporal flicker, spatial coherence, and text–video relevance to provide a fine-grained, multi-dimensional assessment of generated videos.}
  \label{fig:vbench}
\end{figure*}

\section{Datasets}
\label{sec:datasets}
Web-scale corpora such as Koala-36M \cite{koala36m2024}, WebVid-10M \cite{webvid10m2021}, Panda-70M \cite{panda70m2024} and HD-VG-130M \cite{hdvg130m2024} collectively exceed 250 M clips, yet their stock-footage provenance yields noisy captions and licences that forbid commercial use. High-definition, human-centric datasets like CelebV-HQ \cite{celebvhq2022}, OpenHumanVid \cite{openhvid2024}, HumanVid \cite{humanvid2024} and HDTF \cite{hdtf2021} supply face tracks, skeletons and camera-motion labels, but most clips remain under 20 s, limiting long-form training.

Vript \cite{vript2024} provides six-minute films with 145-word scene-level “scripts.” Large-scale video–text datasets like HD-VILA-100M \cite{hd-vila2022} and Panda-70M \cite{panda70m2024} have enabled text-to-video generation at scale, but their clips are mostly 5–15 s with minimal narrative context. To push toward longer, story-driven videos, recent benchmarks offer richer structure: MiraData provides 1–2 min sequences with dense, structured captions covering objects, actions, style and camera motions \cite{miradata2024}, and MovieBench is the first movie-level dataset with hierarchical annotations (movie, scene, shot) enforcing character consistency and multi-scene storytelling \cite{moviebench2025}. Examples of these datasets used in video generation models are in Table~\ref{tab:tasks-datases}.

\section{Evaluation Metrics}
\label{sec:eval}
In recent work, video generation models have largely been assessed using image-derived metrics such as Inception Score (IS) \cite{Salimans2016}, Fréchet Inception Distance (FID) \cite{Heusel2017} and its temporal extension Fréchet Video Distance (FVD) \cite{Unterthiner2019}, along with Structural Similarity Index (SSIM) \cite{wang2004image} and Learned Perceptual Image Patch Similarity (LPIPS) \cite{zhang2018unreasonable}, as well as text alignment via CLIPScore \cite{Hessel2021}. While these measures offer convenient benchmarks, they obscure crucial factors such as temporal coherence, storytelling fidelity and multi-scene consistency, and frequently diverge from human judgments on longer, more complex clips.

To address these shortcomings, VBench introduces a comprehensive, hierarchical evaluation suite that decomposes “video generation quality” into fine-grained dimensions such as visual quality, motion smoothness, identity consistency, temporal flicker, spatial relationships and text-video relevance. Each of these dimensions are driven by tailored prompt sets and validated with human preference annotations \cite{Huang2024}. By providing multi-dimensional scores rather than a monolithic metric, VBench enables detailed diagnostics of generative strengths and weaknesses, making it the principal benchmark for next-generation video models. Another similar benchmark that is multi-dimensional is Wan2.1 \cite{wan2025}, however, VBench has been used by most of the recent literature.

\onecolumn

\addtocounter{table}{-1}

\begin{table*}[!t]
  \centering
  \caption{Overview of video diffusion models, their applications, and training datasets. We categorize models by tasks, generation statistics, subject capabilities, and video duration.
      \xmybox[green]{TV}Text-to-Video, 
    \xmybox[blue]{IV}Image-to-Video, 
    \xmybox[orange]{VV}Video-to-Video, 
    \xmybox[yellow]{VE}Vide-Extension.
    \quad 
    \xmyboxsquare[grass]{S}Single-Subject,
    \xmyboxsquare[mint]{M}Multi-Subject, 
    \xmyboxsquare[bluejeans]{S}Greater than 17 seconds, 
    \xmyboxsquare[yellow]{S}5 up to 16 seconds, 
    \xmyboxsquare[bittersweet]{S}Less than 5 seconds. 
}
\vspace{-3.5mm}
  \label{tab:tasks-datases}
  \small                                  
  \setlength{\tabcolsep}{3.5pt}             
  \begin{tabularx}{\textwidth}{
    p{2.4cm}  |   
    r       |   
    c       |   
    c@{}c@{}c@{}c |  
    c@{\hspace{5pt}}r@{\hspace{0.1pt}}r | 
    c          
    p{2cm}  |   
    p{1.6cm}      
  }
    \toprule

      \bfseries Paper's Github 
      & \bfseries Stars 
      & \bfseries Date
      & \multicolumn{4}{c|}{\bfseries Tasks} 
      & \multicolumn{3}{c|}{\bfseries Generation Statistics}
      & \bfseries Subjects
      & \bfseries Dataset
      & \bfseries Affiliation \\
      \cmidrule(lr){4-7} \cmidrule(lr){9-11}
      & & & \bfseries TV & \bfseries IV & \bfseries VV & \bfseries VE & \bfseries Len. & \bfseries FPS & \bfseries \#Frames & & & \\
    \midrule
    \href{https://seed.bytedance.com/en/seedance}{Seedance 1.0}~\cite{gao2025seedance} & - & 06'25 & \xmybox[green]{TV} & \xmybox[blue]{IV} & & & \xmyboxsquare[yellow]{5} & 30 & 150 & \xmyboxsquare[mint]{M} & - & ByteDance \\
    \href{https://github.com/Tencent-Hunyuan/HunyuanVideo-Avatar}{HunyuanVideo-Avatar}~\cite{chen2025hunyuanvideoavatar} & 1K & 05'25 & \xmybox[green]{TV} & \xmybox[blue]{IV} & & & \xmyboxsquare[bluejeans]{30} & 25 & 750 & \xmyboxsquare[mint]{M} & Koala-36M, CelebV-HQ, HDTF & Tencent \\
    \href{https://github.com/SandAI-org/MAGI-1}{MAGI-1}~\cite{teng2025magi1} & 3.2K & 05'25 & \xmybox[green]{TV} & \xmybox[blue]{IV} & & \xmybox[Yellow]{VE} & \xmyboxsquare[yellow]{16} & 24 & 384 & \xmyboxsquare[grass]{S} & Open-perfectblend & Sand AI \\
    \href{https://github.com/Tencent-Hunyuan/HunyuanCustom}{HunyuanCustom}\\~\cite{hu2025hunyuancustom} & 1K & 05'25 & \xmybox[green]{TV} & \xmybox[blue]{IV} & \xmybox[orange]{VV} & & \xmyboxsquare[yellow]{5} & 26 & 129 & \xmyboxsquare[mint]{M} & OpenHumanvid, Panda-2M & Tencent \\
    \href{https://deepmind.google/models/veo/}{Veo3}~\cite{deepmind2025veo3} & - & 05'25 & \xmybox[green]{TV} & \xmybox[blue]{IV} & & \xmybox[Yellow]{VE} & \xmyboxsquare[yellow]{8} & 60 & 480 & \xmyboxsquare[mint]{M} & - & Google \\
    \href{https://github.com/SkyworkAI/SkyReels-V2}{SkyReels-v2}~\cite{chen2025skyreelsv2} & 2.7K & 04'25 & \xmybox[green]{TV} & \xmybox[blue]{IV} & & \xmybox[Yellow]{VE} & \xmyboxsquare[bluejeans]{30} & 24 & 720 & \xmyboxsquare[mint]{M} & Koala-36M, HumanVid & Skywork AI \\
    \href{https://github.com/hpcaitech/Open-Sora}{Open-Sora 2.0  
    }~\cite{peng2025opensora2} & 26.6K & 03'25 & \xmybox[green]{TV} & \xmybox[blue]{IV} & & & \xmyboxsquare[yellow]{5} & 24 & 128 & \xmyboxsquare[mint]{M} & WebVid-10M, Panda-70M, HD-VG-130M, MiraData, Vript, Inter4K & HPC-AI Tech \\
    \href{https://github.com/Wan-Video/Wan2.1}{WAN}~\cite{wan2025} & 11.8K & 03'25 & \xmybox[green]{TV} & \xmybox[blue]{IV} & & \xmybox[Yellow]{VE} & \xmyboxsquare[yellow]{5} & 16 & 81 & \xmyboxsquare[mint]{M} & - & Alibaba \\
    \href{https://github.com/ali-vilab/VACE}{VACE}~\cite{vace2025} & 2.3K & 03'25 & \xmybox[green]{TV} & \xmybox[blue]{IV} & \xmybox[orange]{VV} & \xmybox[Yellow]{VE} & \xmyboxsquare[yellow]{5} & 16 & 81 & \xmyboxsquare[mint]{M} & - & Alibaba \\
    \href{https://github.com/Phantom-video/Phantom}{Phantom}~\cite{liu2025phantom} & 1K & 02'25 & \xmybox[green]{TV} & \xmybox[blue]{IV} & & & \xmyboxsquare[yellow]{5} & 25 & 125 & \xmyboxsquare[mint]{M} & Panda70M & ByteDance \\
    \href{https://github.com/stepfun-ai/Step-Video-Ti2V}{StepVideo}~\cite{huang2025stepvideo} & 3K & 02'25 & \xmybox[green]{TV} & \xmybox[blue]{IV} & & & \xmyboxsquare[yellow]{8} & 25 & 204 & \xmyboxsquare[mint]{M} & - & Step-Video \\
    \href{https://yuzhou914.github.io/ConceptMaster/}{ConceptMaster}~\cite{huang2025conceptmaster} & - & 01'25 & \xmybox[green]{TV} & \xmybox[blue]{IV} & & & \xmyboxsquare[yellow]{5} & 62 & 310 & \xmyboxsquare[mint]{M} & Panda-2M, MS-COCO & Kuaishou Tech \\
    \href{https://snap-research.github.io/open-set-video-personalization/}{VideoAlchemist}~\cite{chen2025videoalchemist} & - & 01'25 & \xmybox[green]{TV} & \xmybox[blue]{IV} & & & \xmyboxsquare[yellow]{5} & 24 & 120 & \xmyboxsquare[mint]{M} & MSRVTT-Personalization & Snap \\
    \href{https://github.com/Tencent-Hunyuan/HunyuanVideo}{HunyuanVideo}~\cite{kong2024hunyuanvideo} & 10.2K & 12'24 & \xmybox[green]{TV} & \xmybox[blue]{IV} & & \xmybox[Yellow]{VE} & \xmyboxsquare[yellow]{5} & 26 & 129 & \xmyboxsquare[mint]{M} & - & Tencent \\
    \href{https://github.com/Lightricks/LTX-Video}{LTX-video}~\cite{hacohen2025ltxvideo} & 6.3K & 12'24 & \xmybox[green]{TV} & \xmybox[blue]{IV} & & \xmybox[Yellow]{VE} & \xmyboxsquare[yellow]{5} & 24 & 120 & \xmyboxsquare[mint]{M} & MS-COCO, LAION-5B, MSR-VTT, UCF-101 & Lightricks \\
    \href{https://ai.meta.com/research/movie-gen/}{MovieGen}~\cite{polyak2024moviegen} & - & 10'24 & \xmybox[green]{TV} & \xmybox[blue]{IV} & \xmybox[orange]{VV} & & \xmyboxsquare[yellow]{16} & 24 & 384 & \xmyboxsquare[bluejeans]{L} & UCF-101, MSR-VTT, Kinetics-400, SAM-2 & Meta \\
    \href{https://github.com/THUDM/CogVideo}{CogVideoX}~\cite{yang2024cogvideox} & 11.5K & 08'24 & \xmybox[green]{TV} & \xmybox[blue]{IV} & \xmybox[orange]{VV} & & \xmyboxsquare[yellow]{10} & 16 & 160 & \xmyboxsquare[bluejeans]{L} & Panda-70M, COYO-700M, LAION-5B, WebVid & Tsinghua Uni \\
    \href{https://github.com/PKU-YuanGroup/Open-Sora-Plan}{Open-Sora Plan}~\cite{lin2024opensoraplan} & 12K & 11'24 & \xmybox[green]{TV} & \xmybox[blue]{IV} & & & \xmyboxsquare[yellow]{6} & 24 & 144 & \xmyboxsquare[bluejeans]{L} & COCO, JourneyDB, Panda70M, VIDAL-10M, WebVid-10M & Peking Uni \\
  \end{tabularx}
\end{table*}

\addtocounter{table}{-1}

\begin{table*}[!t]\ContinuedFloat
  \caption{(Continued)}
  \small                                  
  \setlength{\tabcolsep}{3.5pt}             
  \begin{tabularx}{\textwidth}{
    p{2.6cm}  |   
    r       |   
    c       |   
    c@{}c@{}c@{}c |  
    c@{\hspace{5pt}}r@{\hspace{0.1pt}}r | 
    c          
    p{2cm}  |   
    p{1.6cm}      
  }
    \toprule
      \bfseries Paper's Github 
      & \bfseries Stars 
      & \bfseries Date
      & \multicolumn{4}{c|}{\bfseries Tasks} 
      & \multicolumn{3}{c|}{\bfseries Generation Statistics}
      & \bfseries Subjects
      & \bfseries Dataset
      & \bfseries Affiliation \\
      \cmidrule(lr){4-7} \cmidrule(lr){9-11}
      & & & \bfseries TV & \bfseries IV & \bfseries VV & \bfseries VE & \bfseries Len. & \bfseries FPS & \bfseries \#Frames & & & \\
    \midrule
    \href{https://yuqingwang1029.github.io/Loong-video/}{Loong}~\cite{wang2024loong} & - & 10'24 & \xmybox[green]{TV} & & & & \xmyboxsquare[bluejeans]{150} & 7 & 1050 & \xmyboxsquare[mint]{M} & LAION-5B, MSR-VTT & ByteDance \\
    \href{https://github.com/Dawn-LX/CausalCache-VDM}{Ca2-VDM}~\cite{gao2024ca2vdm} & 1K & 06'24 & \xmybox[green]{TV} & \xmybox[blue]{IV} & & & \xmyboxsquare[bluejeans]{60} & 24 & 1440 & \xmyboxsquare[grass]{S} & MSR-VTT, UCF-101, Sky Timelapse & Zhejiang Uni \\
    \href{https://openai.com/sora/}{Sora}~\cite{openai2024sora} & - & 06'24 & \xmybox[green]{TV} & \xmybox[blue]{IV} & & & \xmyboxsquare[bluejeans]{20} & 30 & 600 & \xmyboxsquare[mint]{M} & - & openAI \\
    \href{https://github.com/HVision-NKU/StoryDiffusion}{StoryDiffusion}~\cite{zhou2024storydiffusion} & 6.3K & 05'24 & \xmybox[green]{TV} & & & & \xmyboxsquare[yellow]{13} & 14 & 182 & \xmyboxsquare[mint]{M} & Webvid10M & ByteDance \\
    \href{https://github.com/kyegomez/LUMIERE}{Lumiere}~\cite{bartal2024lumiere} & 1K & 01'24 & \xmybox[green]{TV} & \xmybox[blue]{IV} & \xmybox[orange]{VV} & & \xmyboxsquare[yellow]{5} & 16 & 80 & \xmyboxsquare[bluejeans]{L} & UCF101 & Google \\
    \href{https://sites.research.google/videopoet/}{VideoPoet}~\cite{singer2023videopoet} & - & 12'23 & \xmybox[green]{TV} & \xmybox[blue]{IV} & \xmybox[orange]{VV} & & \xmyboxsquare[yellow]{5} & 8 & 41 & \xmyboxsquare[bluejeans]{L} & MSR-VTT, UCF-101, Kinetics 600, Something-Something V2, DAVIS & Google \\
    \href{https://github.com/ali-vilab/Vgen}{DreamVideo}~\cite{gao2023dreamvideo} & 3.1K & 12'23 & \xmybox[green]{TV} & \xmybox[blue]{IV} & \xmybox[orange]{VV} & & \xmyboxsquare[bittersweet]{4} & 8 & 32 & \xmyboxsquare[grass]{S} & UCF101, DAVIS & Alibaba \\
    \href{https://github.com/ali-vilab/Vgen}{I2VGen-XL}~\cite{zhang2023i2vgenxl} & 3.1K & 11'23 & \xmybox[green]{TV} & \xmybox[blue]{IV} & \xmybox[orange]{VV} & & \xmyboxsquare[bluejeans]{30} & 8 & 200 & \xmyboxsquare[grass]{S} & Web-Vid10M, LAION-400M & Alibaba \\
    \href{https://github.com/Vchitect/SEINE}{SEINE}~\cite{chen2023seine} & 1K & 11'23 & \xmybox[green]{TV} & & & & \xmyboxsquare[bittersweet]{2} & 8 & 16 & \xmyboxsquare[grass]{S} & UCF101 & Shanghai AI \\
    \href{https://github.com/Stability-AI/generative-models}{SVD}~\cite{ho2023svd} & 25.9K & 11'23 & \xmybox[green]{TV} & \xmybox[blue]{IV} & & & \xmyboxsquare[bittersweet]{2} & 14 & 25 & \xmyboxsquare[grass]{S} & UCF101, LVD-10M, MVImgNet, Google Scanned Objects & Stability AI \\
    \href{https://magvit.cs.cmu.edu/v2/}{MAGVIT-v2}~\cite{yu2023magvitv2} & - & 10'23 & \xmybox[green]{TV} & \xmybox[blue]{IV} & & & \xmyboxsquare[bittersweet]{2} & 8 & 17 & \xmyboxsquare[grass]{S} & UCF-101, Kinetics-600, SSv2 & Google \\
    \href{https://github.com/guoyww/AnimateDiff}{AnimateDiff}~\cite{gu2023animatediff} & 11.4K & 07'23 & & \xmybox[blue]{IV} & & & \xmyboxsquare[bittersweet]{2} & 16 & 32 & \xmyboxsquare[grass]{S} & WebVid-10M, Civitai & Stanford Uni \\
    \href{https://video-diffusion.github.io/}{VDM}~\cite{ho2022vdm} & - & 04'22 & \xmybox[green]{TV} & \xmybox[blue]{IV} & & & \xmyboxsquare[bittersweet]{2} & 8 & 16 & \xmyboxsquare[grass]{S} & UCF101, BAIR Robot Pushing, Kinetics-600 & Google \\
  \end{tabularx}
\end{table*}

  \vspace{1ex}

\twocolumn

\end{document}